\documentclass{article}

\PassOptionsToPackage{numbers, compress}{natbib}
\usepackage[preprint]{main}

\usepackage[utf8]{inputenc}
\usepackage[T1]{fontenc}
\usepackage{hyperref}
\usepackage{url}
\usepackage{booktabs}
\usepackage{xcolor}
\usepackage{colortbl}
\usepackage{amsfonts}
\usepackage{nicefrac}
\usepackage{microtype}
\usepackage{graphicx}
\usepackage{multirow}
\usepackage{amsmath}
\usepackage{algorithm}
\usepackage{algpseudocode}
\usepackage{float}
\usepackage{placeins}
\usepackage{lineno}

\newcommand{\ColGraphRAG}{ColGraphRAG}

\title{ColGraphRAG: Late-Interaction Evidence Retrieval\\for Multimodal GraphRAG}

\author{
  Seonok Kim \\
  Mazelone \\
  \texttt{seonokrkim@gmail.com}
}

\begin{document}

\maketitle
\raggedbottom

\begin{abstract}
Graph-grounded multimodal question answering organizes text, tables, and images in a structured evidence graph, yet end-to-end accuracy depends on which multimodal assets are ranked highly enough to enter downstream reasoning; for graph-linked images, single-vector bi-encoder similarity can discard patch- and token-level structure needed for fine-grained alignment.
We evaluate replacing the visual candidate-ranking operator over graph-linked image nodes with late-interaction MaxSim-style multi-vector scoring in the ColBERT/ColPali lineage, while keeping offline graph construction, text- and table-side retrieval, structured extraction, and downstream reasoning unchanged.
On MultimodalQA, this change is associated with improved retrieval-stage point estimates for graph-linked image candidates and downstream QA gains, with larger movement where visual evidence matters most and mixed trends on text-dominant questions; we interpret the pattern as mechanism-level evidence for graph-linked visual evidence inclusion, while broader validation and finer graph-level diagnostics remain important future work.
\end{abstract}

\section{Introduction}
\label{sec:intro}

Multimodal question answering over long, visually rich documents requires evidence from text, tables, and figures that may only become decisive when combined.
Graph-grounded systems link assets in an evidence graph and reason over a subgraph, yet downstream stages only see evidence promoted early enough; decisive figures or tables that rank too low are often irrecoverable.
Recent long-document and visually grounded QA benchmarks reinforce this setting, where evidence is scattered across pages and modalities~\citep{ma2024mmlongbenchdoc,deng2025longdocurl,dong2025mmdocir,wasserman2025real,jain2025simpledoc,chia2025mlongdoc,vanlandeghem2023dude,blau2024gram}.

A recurring weak point is the \emph{ranking of graph-linked image candidates before evidence assembly}.
Query-driven multimodal GraphRAG-style stacks often still rank images with CLIP-style bi-encoders~\cite{radford2021clip} that map each query--image pair to one similarity score per side, pooling away local structure that matters when questions refer to specific regions or fine visual cues.
Fine-grained late-interaction multimodal retrievers improve visually grounded retrieval in RA-VQA~\cite{lin2023flmr,lin2024preflmr}, complementing pooled bi-encoder design.

We study the retrieval-to-graph bottleneck by replacing the graph-linked visual ranker within the same GraphRAG pipeline.
At inference, ColGraphRAG operates on the same pre-built graph $G$ as the pooled-ranking baseline but replaces the graph-linked visual ranker with ColBERT/ColPali-style MaxSim scoring over multiple query vectors and local visual units, using deterministic template-based visual query phrases; offline construction and non-visual retrieval match the baseline.

On MultimodalQA, replacing pooled visual ranking is associated with improved retrieval-stage point estimates on graph-linked image candidates and higher aggregate EM/F1 point estimates relative to the same system with single-vector ranking; modality-level patterns include text-dominant subsets where scores need not move favorably.
We read these results as mechanism-level evidence that finer-grained visual ranking can improve graph-linked evidence inclusion, without claiming full-benchmark superiority or statistical significance.

Using the same $G$ and non-visual retrieval lets us ask whether finer query--image matching improves the evidence available to downstream graph reasoning.

\paragraph{Contributions.}
\begin{itemize}
\item We formulate graph-linked image candidate ranking as an evidence-inclusion
bottleneck in multimodal GraphRAG: relevant visual nodes may exist in the graph
but fail to enter structured extraction if pooled ranking suppresses them.

\item We introduce ColGraphRAG, which replaces pooled single-vector visual
candidate ranking with late-interaction MaxSim scoring over graph-linked image
candidates while reusing the same graph, non-visual retrieval, extraction, and
answering stack.

\item We evaluate the retrieval-to-graph effect across candidate-list metrics,
end-task QA scores, and modality breakdowns on MultimodalQA, with WebQA and
ViDoRe v3 serving as contextual and retrieval-native evaluations.
\end{itemize}

\section{Related work}
\label{sec:related_work}

\subsection{Multimodal QA over long and visually rich documents}
Multimodal question answering studies how to answer questions requiring evidence from heterogeneous sources such as text, tables, and images~\cite{talmor2021multimodalqa}.
In long or multi-page settings, relevant evidence is often dispersed across passages, structured tables, and visually rich content, making coordinated retrieval across modalities essential~\cite{suri2025visdom,tanaka2025vdocrag}.
Recent long-document benchmarks explicitly stress multi-page understanding and visually grounded reading~\cite{ma2024mmlongbenchdoc,deng2025longdocurl}; retrieval-oriented resources further characterize long multimodal documents~\cite{dong2025mmdocir}, realistic multimodal RAG settings~\cite{wasserman2025real}, and multi-page document QA~\cite{vanlandeghem2023dude,blau2024gram,jain2025simpledoc,chia2025mlongdoc}.
These benchmarks motivate our setting: evidence is heterogeneous, but downstream reasoning only sees assets promoted by retrieval.
We therefore study graph-linked visual evidence routing in MultimodalQA within a graph-grounded QA pipeline where the ranking interface is the only changed component.

\subsection{Graph-grounded or graph-modeled multimodal retrieval and reasoning}
Multi-hop QA benchmarks such as HotpotQA~\cite{yang2018hotpotqa} and MuSiQue~\cite{trivedi2022musique} stress structured, multi-step evidence assembly.
Multimodal graph encoders, query-conditioned graph construction, and layout-aware graph modeling further organize heterogeneous evidence for retrieval-augmented reasoning~\cite{he2023multimodalgraphtransformer,bu2025querydriven,yang-etal-2025-superrag}.
Those lines primarily reshape how graphs are built or traversed; ColGraphRAG targets the retrieval-to-graph interface that decides which graph-linked visual candidates structured extraction and downstream reasoning can use.

\subsection{Late-interaction retrieval and local visual grounding}
Single-vector bi-encoder retrieval is efficient but can miss fine-grained matches between query concepts and local visual evidence~\cite{radford2021clip}.
Late-interaction models preserve multiple query and document vectors and aggregate local matches with MaxSim-style scoring~\cite{khattab2020colbert,santhanam2022colbertv2}.
Multimodal extensions study fine-grained late interaction for visually grounded retrieval in RA-VQA~\cite{lin2023flmr,lin2024preflmr}.
Vision--language document retrieval increasingly uses patch- or token-level representations for pages and documents~\cite{faysse2025colpali}.
These methods motivate our scorer choice.
We apply MaxSim directly to graph-linked image candidates induced by the upstream evidence graph, placing fine-grained visual matching at the retrieval-to-graph boundary where ranking decides which image evidence enters downstream graph stages.
We do not propose a new graph constructor; we isolate the scorer at the retrieval-to-graph boundary while using the same graph-grounded stack.
We follow the ColBERT/ColPali multi-vector MaxSim \emph{interface}~\cite{khattab2020colbert,faysse2025colpali}; Section~\ref{sec:method} formalizes the scorer used in ColGraphRAG.

\section{Method}
\label{sec:method}

\begin{figure*}[t]
  \centering
  \includegraphics[width=\textwidth]{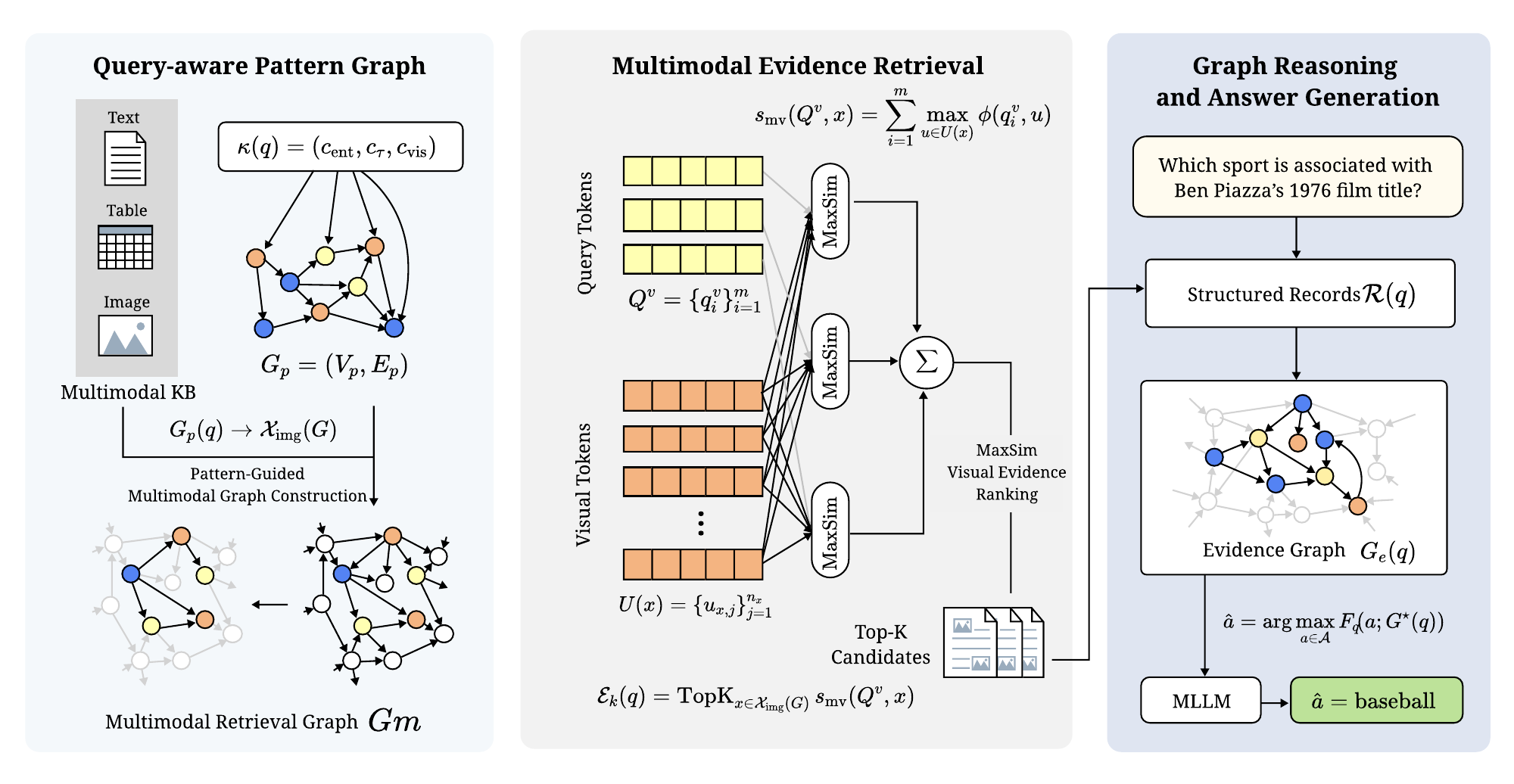}
  \caption{Overview of ColGraphRAG.
Given a multimodal knowledge base, ColGraphRAG derives a query-aware pattern graph and forms a multimodal retrieval graph using text, table, and image evidence.
The figure uses $G_p$ and $G_m$ as compact visual labels; in the method, these objects are written as $G_p(q)$ and $G_m(q)$ to emphasize query conditioning.
The visual branch ranks graph-linked image candidates $x\in\mathcal{X}_{\mathrm{img}}(G)$.
Query-side tokens $Q^v=\{q_i^v\}_{i=1}^{m}$ encode deterministic visual phrases $C_{\mathrm{vis}}(q)=\mathrm{ImagePhrases}(q,\kappa(q),G)$, while each candidate exposes local visual units $U(x)=\{u_j\}_{j=1}^{n_x}$.
Late-interaction MaxSim produces an ordered top-$k$ graph-linked image candidate list $E_k(q)$ (Sec.~\ref{sec:method:late}).
Retrieved evidence is assembled into $G_e(q)$, expanded into the reasoning graph $G^{\star}(q)$, and serialized for answer generation.}
  \label{fig:overview}
\end{figure*}

We describe \ColGraphRAG{} as a modification of a graph-grounded multimodal QA pipeline: everything except the visual candidate-ranking operator over graph-linked image nodes is unchanged---pooled single-vector similarity is replaced with late-interaction MaxSim-style multi-vector scoring.
Figure~\ref{fig:overview} sketches $G_p(q) \rightarrow$ graph-linked visual ranking over $\mathcal{X}_{\mathrm{img}}(G)$ $\rightarrow G_m(q) \rightarrow G_e(q) \rightarrow G^{\star}(q)$.
Throughout, graph-linked candidates are $x$ with local visual units $U(x)$.

\subsection{Problem setup}
\label{sec:method:setup}

Let $q$ denote a natural-language question.
For structured retrieval and extraction we summarize clues by
\[
  \kappa(q)=(c_{\mathrm{ent}},c_\tau,c_{\mathrm{vis}}).
\]
Here $c_{\mathrm{ent}}$ is an entity clue, $c_\tau$ is a temporal clue, and $c_{\mathrm{vis}}$ is a visual clue phrase feeding template-based phrase construction for the image branch.
Deterministic visual phrases are
\[
  C_{\mathrm{vis}}(q)=\mathrm{ImagePhrases}(q,\kappa(q),G).
\]
This yields a finite phrase set induced from $\kappa(q)$ under fixed templates on $G$---not an additional generative VLM output for ranking.
Visual query tokens for late interaction are
\[
  Q^v=\{q_i^v\}_{i=1}^{m}=\mathrm{Enc}(C_{\mathrm{vis}}(q)).
\]
Let $\mathcal{X}$ denote a finite multimodal candidate pool (pages, passages, images, tables) attached to $G$.
We distinguish the full pool $\mathcal{X}$ from the graph-linked image candidate pool $\mathcal{X}_{\mathrm{img}}(G)$, which is the only pool rescored by late interaction in this work.
The distinction matters: late interaction only reranks the graph-linked image pool, while text and table retrieval remain unchanged.
For the \emph{visual} branch, each image-linked candidate $x\in\mathcal{X}_{\mathrm{img}}(G)$ is represented by local units
\[
  U(x)=\{u_j\}_{j=1}^{n_x}\,,
\]
where each $u_j$ may be a patch, region, or other visual fragment.
The goal is to predict an answer by retrieving a relevant evidence subset, grounding it in an evidence graph, and reasoning over that graph.

The single-vector \emph{visual} baseline pools query and image to one embedding each before matching; ColGraphRAG replaces only that operator with \emph{late-interaction MaxSim} scoring over $\mathcal{X}_{\mathrm{img}}(G)$ so alignment stays fine-grained until scoring.
Text/table retrieval follows the baseline; promoted evidence still gates what downstream stages can use.
When only visual ranking changes while $G$, $B(q)$, extraction templates, and the generator are unchanged, end-to-end shifts trace to the visual candidate-ranking interface---useful attribution for an isolatable bottleneck.

\subsection{Offline construction and inference on a pre-built graph}
\label{sec:method:pipeline}

\noindent
\begin{minipage}[t]{0.49\textwidth}
  \vspace{0pt}
  \noindent\textbf{Offline construction.}
  We separate \emph{offline} graph construction from \emph{online} inference.
  During offline construction, the multimodal corpus is parsed into text, table, image, and entity nodes and serialized (e.g., GraphML) as a pre-built graph $G$.

  \medskip
  \noindent\textbf{Inference on a pre-built graph.}
  Algorithm~\ref{alg:colgraphrag-inference} summarizes the visual-ranker comparison: graph construction, baseline non-visual retrieval, and downstream graph-grounded answering are unchanged, while only the operator that ranks graph-linked image candidates is replaced with late-interaction MaxSim scoring.
  At inference time, the user query is processed against this graph: baseline text/table/entity retrieval $R_{\mathrm{base}}$ is unchanged, while the image branch applies $R_{\mathrm{img}}$ over graph-linked \textsc{IMAGE} nodes.
  The image branch forms deterministic template-based visual query phrases and ranks candidates with late-interaction MaxSim scoring on $G$.
  This phrase construction is \emph{deterministic and does not call a generative LLM} in the evaluated pipeline; it only supplies text tokens for alignment with image encoders.
  Extraction yields $G_e(q)$, which is expanded on $G$ to $G^{\star}(q)$ for reasoning and answer generation.
\end{minipage}%
\hfill
\begin{minipage}[t]{0.48\textwidth}
  \vspace{0pt}
  \centering
  \small
  \vspace{-1.5em}
  \begin{algorithm}[H]
    \raggedright\small
    \caption{ColGraphRAG Inference with Late-Interaction Candidate Ranking}
    \label{alg:colgraphrag-inference}

\begin{algorithmic}[1]
\Require Question $q$, pre-constructed graph $G$, candidate budget $k$, expansion rounds $T$
\Require Baseline non-visual retrieval $R_{\mathrm{base}}$; late-interaction image ranking $R_{\mathrm{img}}$
\Ensure Predicted answer $\hat{a}$.

\State $B(q) \leftarrow R_{\mathrm{base}}(q, G)$
\Comment{text/table/entity retrieval on nodes in $G$}

\State $P(q) \leftarrow \mathrm{BuildPattern}(q,\kappa(q),G)$
\Comment{query-conditioned extraction pattern}

\State $C_{\mathrm{vis}}(q) \leftarrow \mathrm{ImagePhrases}(q,\kappa(q),G)$
\Comment{deterministic template-based visual query phrases}

\State $E_k(q) \leftarrow R_{\mathrm{img}}\bigl(C_{\mathrm{vis}}(q), X_{\mathrm{img}}(G), k\bigr) \quad \triangleright\ \text{ordered top-$k$ graph-linked image candidates}$

\State $R(q) \leftarrow \mathrm{Extract}\bigl(E_k(q),B(q),P(q)\bigr)$
\State $G_e(q) \leftarrow \mathrm{AssembleGraph}(R(q))$
\State $G^{\star}(q) \leftarrow G_e(q)$

\For{$t = 1$ to $T$}
    \If{$\mathrm{Sufficient}(q, G^{\star}(q))$}
        \State \textbf{break}
    \EndIf
    \State $G^{\star}(q) \leftarrow \mathrm{ExpandOnGraph}(G^{\star}(q), G)$
\EndFor

\State $z_{\mathrm{ser}} \leftarrow \mathrm{Serialize}(G^{\star}(q))$
\State $\hat{a} \leftarrow \mathrm{Generate}(q, z_{\mathrm{ser}})$
\State \Return $\hat{a}$
\end{algorithmic}

  \end{algorithm}
\end{minipage}

\par\medskip
\noindent
All optional augmentation modules are disabled in MultimodalQA runs; both systems share $k$, $T$, the graph snapshot, non-visual retrievers, extraction templates, and generator.

\subsection{Late-interaction visual retrieval}
\label{sec:method:late}
\label{sec:method:compare}

\begin{figure*}[!t]
  \centering
  \includegraphics[width=\textwidth]{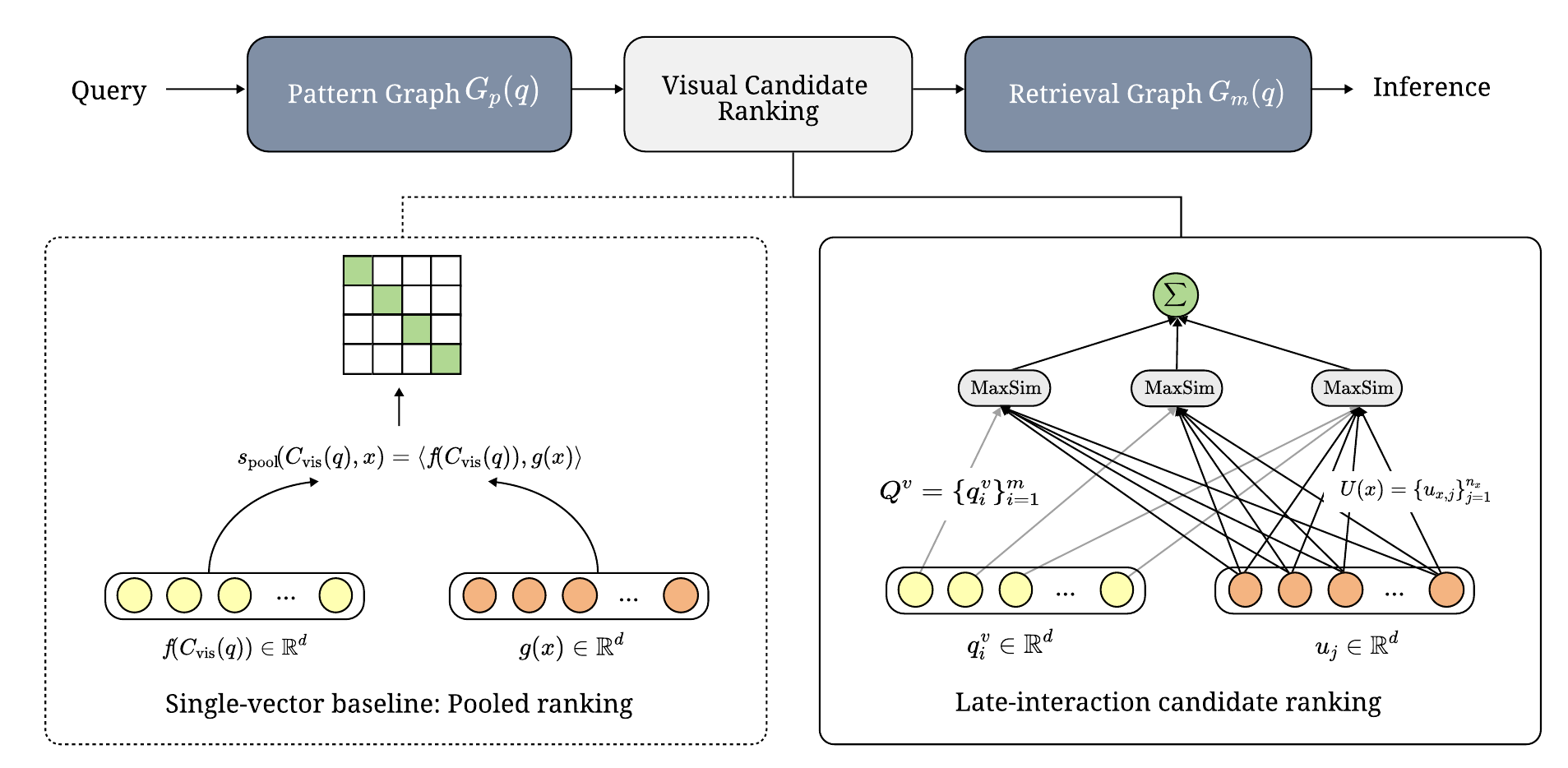}
  \caption{Single-vector pooled ranking versus late-interaction MaxSim candidate ranking.
  Both variants follow the same route $q \rightarrow G_p(q) \rightarrow$ visual candidate ranking $\rightarrow G_m(q) \rightarrow$ inference and rank the same fixed graph-linked image candidate pool $\mathcal{X}_{\mathrm{img}}(G)$.
  The pooled baseline maps deterministic visual phrases $C_{\mathrm{vis}}(q)$ and each candidate $x$ to one vector per side and scores them by $s_{\mathrm{pool}}(C_{\mathrm{vis}}(q),x)=\langle f(C_{\mathrm{vis}}(q)),g(x)\rangle$.
  The late-interaction ranker instead preserves query-side token vectors $Q^v=\{q_i^v\}_{i=1}^{m}$ and local visual units $U(x)=\{u_j\}_{j=1}^{n_x}$, both in $\mathbb{R}^d$, and scores candidates by $s_{\mathrm{mv}}(Q^v,x)=\sum_{i=1}^{m}\max_{u\in U(x)}\phi(q_i^v,u)$.
  Only this graph-linked visual ranking operator is substituted.}
  \label{fig:comparison}
\end{figure*}

Figure~\ref{fig:comparison} situates the swap along the shared route $q \rightarrow G_p(q) \rightarrow$ \emph{graph-linked visual candidate ranking over $\mathcal{X}_{\mathrm{img}}(G)$} $\rightarrow G_m(q) \rightarrow$ inference, then contrasts (\textit{i})~pooled single-vector scoring with (\textit{ii})~late-interaction MaxSim over $Q^v$ and $U(x)$.
Both sides rank the same fixed pool $\mathcal{X}_{\mathrm{img}}(G)$; only the scoring operator differs.
The pooled single-vector baseline maps deterministic visual phrases to one embedding per side:
\[
  s_{\mathrm{pool}}(C_{\mathrm{vis}}(q),x)
  =
  \langle f(C_{\mathrm{vis}}(q)),g(x)\rangle\,,
  \qquad f(C_{\mathrm{vis}}(q)),g(x)\in\mathbb{R}^d\,.
\]
The proposed operator instead keeps token vectors $Q^v=\mathrm{Enc}(C_{\mathrm{vis}}(q))$ and local candidate units $U(x)$, scoring with a MaxSim-style aggregate.

For each graph-linked image candidate $x\in\mathcal{X}_{\mathrm{img}}(G)$, query-side tokens $Q^v=\{q_i^v\}_{i=1}^{m}$ and local units $U(x)=\{u_j\}_{j=1}^{n_x}$ lie in the same retrieval space.
The graph-linked visual \emph{candidate-ranking} score is
\begin{equation}
  s_{\mathrm{mv}}(Q^v,x)
  =
  \sum_{i=1}^{m}
  \max_{u\in U(x)}
  \phi(q_i^v,u)\,,
  \label{eq:maxsim}
\end{equation}
where $\phi(\cdot,\cdot)$ denotes cosine similarity or dot product after $\ell_2$ normalization, depending on the encoder.
This differs from pooled retrieval, which collapses each side to one vector before matching.
Deferring pooling preserves local evidence when answers hinge on small visual cues---common when graph-linked nodes are whole pages or figures but the query targets a localized object, label, or region that pooled scores can miss.
We refer to Eq.~\ref{eq:maxsim} as \emph{MaxSim-style late-interaction} scoring in the ColBERT/ColPali lineage~\cite{khattab2020colbert,faysse2025colpali}; it is \emph{not} applied to text or table rows in our pipeline.
Let $\mathcal{X}_{\mathrm{img}}(G)$ denote graph-linked image candidates on $G$.
We define the top-$k$ graph-linked image evidence list by late-interaction score as
\[
  E_k(q)=
  \operatorname{TopK}^{\mathrm{ranked}}_{x\in\mathcal{X}_{\mathrm{img}}(G)}
  s_{\mathrm{mv}}(Q^v,x),
\]
where $E_k(q)$ denotes the ordered top-$k$ graph-linked image candidate list sorted by $s_{\mathrm{mv}}$.
Retrieval-stage metrics such as Hit@1, MRR@$k$, and nDCG@$k$ are computed on this ranked order, while downstream extraction consumes the top-$k$ members.
Ties are broken arbitrarily and $k$ is held constant in experiments.
Thus $E_k(q)$ draws only from $\mathcal{X}_{\mathrm{img}}(G)\subseteq\mathcal{X}$; we never rank over the full multimodal pool $\mathcal{X}$ for this visual gate.
Baseline text/table retrieval outputs are $B(q)$; together with $E_k(q)$ they feed structured extraction.

\begin{figure*}[t]
  \centering
  \includegraphics[width=\textwidth]{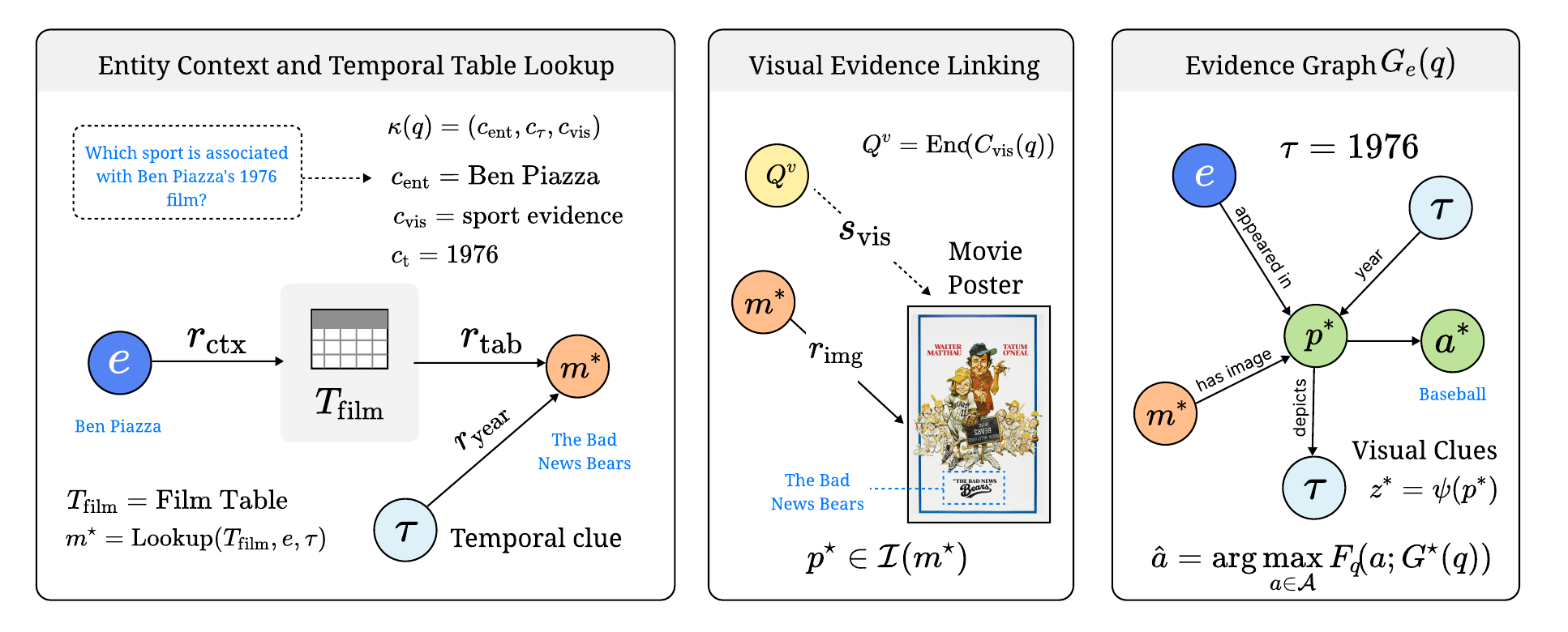}
  \caption{Illustrative case study of evidence linking and graph-based reasoning.
For the query asking which sport is associated with Ben Piazza's 1976 film, $\kappa(q)=(c_{\mathrm{ent}},c_\tau,c_{\mathrm{vis}})$ summarizes entity, temporal, and visual clues.
The temporal clue $c_\tau=1976$ guides a table lookup to a film entity $m^\star$, and the visual branch retrieves an associated poster candidate $p^\star\in\mathcal{I}(m^\star)$.
The selected visual evidence is converted into a symbolic visual clue $z^\star=\psi(p^\star)$ and inserted into $G_e(q)$; expansion yields $G^{\star}(q)$ for answer generation.
Edges $r_{\mathrm{ctx}}$, $r_{\mathrm{tab}}$, $r_{\mathrm{year}}$, and $r_{\mathrm{img}}$ denote context, table lookup, temporal, and image-linking relations used only to illustrate the evidence path.}
  \label{fig:casestudy}
\end{figure*}

\subsection{Query-conditioned structured extraction}
\label{sec:method:extract}

Given the ordered top-$k$ graph-linked image candidate list $E_k(q)$ \emph{and} text- and table-linked nodes retrieved from $G$ by the baseline graph-grounded procedure (not rescored by late interaction), \ColGraphRAG{} switches from dense scoring to structured evidence processing.
We derive a query-conditioned extraction pattern $P(q)$ that specifies a small schema of entity types, relation labels, and table fields to inspect; extraction is constrained to retrieved evidence and preserves source provenance.
In our evaluated pipeline, this step does not rescore candidates and does not introduce additional visual retrieval.
We run constrained extraction on the combined evidence bundle:
\[
  R(q)=\mathrm{Extract}\bigl(E_k(q),B(q),P(q)\bigr).
\]
Here $B(q)$ denotes baseline text- and table-side retrieval outputs from $G$.
$R(q)$ is a set of structured records (entity tuples, relation tuples, optional table-linked rows).

Extraction is query-aware and limited to retrieved candidates (no full-corpus parse at inference), keeping structured outputs compact and retrieval-tied.

\subsection{Structured evidence graph construction}
\label{sec:method:graph}

We map $R(q)$ to a heterogeneous structured evidence graph $G_e(q)=(V_e(q),A_e(q))$.
Nodes $V_e(q)$ include normalized entities, table rows or table anchors, graph-linked image evidence units, and provenance-bearing source nodes; arcs $A_e(q)$ store extracted relations plus provenance links back to sources.
Construction is deterministic: records are parsed, normalized, merged by identity, and linked with rule-based assembly rather than latent edge prediction.

\ColGraphRAG{} applies four steps:
(1) \emph{Entity normalization and merge} to unify co-referring mentions;
(2) \emph{Relation insertion and aggregation} to add or strengthen edges when multiple records support the same relation;
(3) \emph{Table attachment}, representing tables as structured nodes linked to anchor entities; and
(4) \emph{Provenance preservation}, recording document, page, image, or table identifiers for each node and edge.

The graph is \emph{not} a vector-valued memory: it is a structured, interpretable substrate that organizes evidence already chosen by the retriever.


\FloatBarrier

\subsection{Graph-grounded answering}
\label{sec:method:answer}

Evidence assembly yields $G_e(q)$; Algorithm~\ref{alg:colgraphrag-inference} initializes $G^{\star}(q)\leftarrow G_e(q)$ and applies $\mathrm{ExpandOnGraph}(\cdot,G)$ for up to $T$ rounds until a sufficiency check passes.
Compactly,
\begin{equation}
  G^{\star}(q)=\mathrm{ExpandOnGraph}(G_e(q),G,T)\,,
  \qquad
  \hat{a}=\mathrm{Generate}\bigl(q,\mathrm{Serialize}(G^{\star}(q))\bigr)\,,
\end{equation}
where $\mathrm{ExpandOnGraph}(G_e(q),G,T)$ denotes that iterative expansion contract (rather than a single-hop rewrite).
Equivalently, writing $F_q$ for the induced scoring functional over answer strings given serialized graph context,
\begin{equation}
  \hat{a}=\arg\max_{a\in\mathcal{A}} F_q\bigl(a;G^{\star}(q)\bigr)\,.
\end{equation}
We attach neighboring entities, relations, and table-linked nodes during expansion to form a compact answering subgraph, serialized to structured text for the generator.

Retrieval proposes evidence, graph construction organizes it, generation composes the answer; in this regime graph-grounded QA can bottleneck on \emph{evidence inclusion}, since downstream stages cannot invent missing image support without promotion into the evidence assembly set.
Late interaction keeps fine-grained alignment until scoring and mainly targets \emph{candidate omission}---the node exists in $G$ but pooled ranking excludes it from $E_k(q)$---rather than graph-missing assets or extraction and reasoning failures once evidence is present; empirical results below interpret along this chain.

\section{Experiments}
\label{sec:exp}

%

\begin{table}[t]
  \centering
  \caption{Contextual end-task results on MultimodalQA and WebQA.
  MultimodalQA reports F1 and exact match (EM); WebQA reports \textbf{QA-FL}, \textbf{QA-Acc}, and corpus-level \textbf{QA}.
  Upper rows reproduce protocol-specific literature numbers and are included only to contextualize scale; they are not treated as an apples-to-apples leaderboard.
  The bottom two rows compare QD-MMGraphRAG and ColGraphRAG under the same graph snapshot, candidate budget, non-visual retrieval, extraction interface, expansion behavior, and generator, differing only in the visual candidate-ranking operator.
  Within each Type block and metric column, bold marks the highest displayed value and underline marks the second-highest displayed value; em dashes are excluded.}
  \label{tab:mmqa-webqa-combined}
  \small
  \setlength{\tabcolsep}{3pt}
  \resizebox{\textwidth}{!}{%
  \begin{tabular}{@{}c | c | cc | cc | cc | c | ccc@{}}
    \toprule
    \multirow{3}{*}{\textbf{Type}}
      & \multicolumn{7}{c|}{\textbf{MultimodalQA}} & \multicolumn{4}{c}{\textbf{WebQA}} \\
    \cmidrule(lr){2-8} \cmidrule(l){9-12}
    & \multirow{2}{*}{\textbf{Model}}
      & \multicolumn{2}{c}{\textbf{Unimodal}} & \multicolumn{2}{c}{\textbf{Multimodal}} & \multicolumn{2}{c|}{\textbf{All}}
      & \multirow{2}{*}{\textbf{Model}} & \multirow{2}{*}{\textbf{QA-FL}} & \multirow{2}{*}{\textbf{QA-Acc}} & \multirow{2}{*}{\textbf{QA}} \\
    \cmidrule(lr){3-4} \cmidrule(lr){5-6} \cmidrule(lr){7-8}
    & & \textbf{F1} & \textbf{EM} & \textbf{F1} & \textbf{EM} & \textbf{F1} & \textbf{EM} & & & & \\
    \midrule
    \multirow{4}{*}{Supervised}
      & AR~\citep{talmor2021multimodalqa} & 58.5 & 51.7 & 40.2 & 34.2 & 51.1 & 44.7 & VLP-x101fpn~\citep{chang2022webqa} & 42.6 & 36.7 & 22.6 \\
     & ID~\citep{talmor2021multimodalqa} & 58.4 & 51.6 & 51.2 & 44.6 & 55.5 & 48.8 & VLP-VinVL~\citep{chang2022webqa} & 44.2 & 38.9 & 24.1 \\
     & MMHQA-ICL~\citep{liu2023mmhqa_icl} & \textbf{72.9} & \underline{60.5} & \underline{55.5} & \underline{46.2} & \textbf{65.8} & \underline{54.8} & MuRAG~\citep{chen2022murag} & \textbf{55.7} & \underline{54.6} & \underline{36.1} \\
     & SKURG~\citep{yang2023enhancing_multimodal_multihop} & \underline{69.7} & \textbf{66.1} & \textbf{57.2} & \textbf{52.5} & \underline{64.0} & \textbf{59.8} & SKURG~\citep{yang2023enhancing_multimodal_multihop} & \underline{55.4} & \textbf{57.1} & \textbf{37.7} \\
    \midrule
    \multirow{7}{*}{Unsupervised}
      & Vicuna-7B~\citep{chiang2023vicuna} & 20.3 & 17.1 & 16.4 & 11.9 & 18.6 & 14.9 & Vicuna-7B~\citep{chiang2023vicuna} & 20.6 & 38.1 & 12.1 \\
     & Llama2Chat-13b~\citep{touvron2023llama2} & 24.6 & 21.3 & 17.3 & 13.0 & 21.5 & 17.7 & Llama2Chat-13b~\citep{touvron2023llama2} & 19.5 & 38.0 & 13.2 \\
     & OpenChat-v2-w-13b~\citep{wang2024openchat} & 25.3 & 22.0 & 18.9 & 15.5 & 22.5 & 19.2 & MOQAGPT~\citep{zhang2023moqagpt} & 25.5 & 44.8 & 17.1 \\
     & MOQAGPT~\citep{zhang2023moqagpt} & 54.6 & 49.1 & 33.8 & 30.5 & 45.6 & 41.1 & OFA-Cap~\citep{wang2022ofa} & 52.8 & 55.4 & 33.5 \\
     & Binder~\citep{cheng2023binding} & --- & --- & --- & --- & \underline{57.1} & 51.0 & PROMPTCap~\citep{hu2023promptcap} & 53.0 & 57.2 & 34.5 \\
     & QD-MMGraphRAG~\citep{bu2025querydriven} & \underline{66.4} & \underline{64.5} & \underline{35.5} & \underline{34.2} & 54.7 & \underline{53.0} & QD-MMGraphRAG~\citep{bu2025querydriven} & \underline{58.4} & \underline{63.1} & \underline{43.6} \\
     & ColGraphRAG (Ours) & \textbf{68.6} & \textbf{65.5} & \textbf{45.6} & \textbf{42.2} & \textbf{58.3} & \textbf{55.0} & ColGraphRAG (Ours) & \textbf{77.1} & \textbf{73.6} & \textbf{58.4} \\
    \bottomrule
  \end{tabular}%
  }
\end{table}

\subsection{Experimental objective and design}
\label{sec:exp:objective}

We test whether late-interaction MaxSim changes which graph-linked image evidence reaches $E_k(q)$ early enough to affect downstream reasoning and answering when visuals matter.
Only that candidate-ranking operator swaps from pooled similarity to MaxSim; $G$, $B(q)$, extraction, evidence-graph construction, and generation are shared, attributing differences to the retrieval-to-graph gateway rather than a new stack.

The intended causal chain is:
\[
  \mathrm{MaxSim}
  \Rightarrow
  E_k(q)
  \Rightarrow
  R(q)
  =
  \mathrm{Extract}\bigl(E_k(q), B(q), P(q)\bigr)
  \Rightarrow
  G^{\star}(q)
  \Rightarrow
  \hat{a}.
\]
Here $E_k(q)$ is the ordered top-$k$ graph-linked image candidate list, $R(q)$ is the structured record bundle, and $G^{\star}(q)$ is the question-specific reasoning subgraph used for generation.
We do \emph{not} treat aggregate leaderboard rank as the scientific target; end-task metrics support the downstream link in this chain under the same downstream machinery.

\subsection{Datasets and evaluation protocol}
\label{sec:exp:datasets}

We evaluate the same visual-ranker substitution on graph-grounded QA, a contextual end-task benchmark, and retrieval-native ViDoRe v3.

\textbf{MultimodalQA} \citep{talmor2021multimodalqa} is primary.
Questions may blend text, tables, and images; we test whether MaxSim changes (i)~membership in $E_k(q)$ and (ii)~EM/F1 after unchanged extraction/generation.
Modality tags reflect dominant gold evidence type and do not isolate pure single-modality runs when assembly still mixes assets.

\textbf{WebQA} \citep{chang2022webqa} appears as a paired block in Table~\ref{tab:mmqa-webqa-combined}.
Upper literature rows report numbers under each source paper's protocol and are not treated here as strict apples-to-apples reimplementations in our stack.
The \textbf{bottom two rows} compare the two GraphRAG variants under the same graph snapshot, candidate budget, non-visual retrieval, extraction interface, expansion behavior, and generator, differing only in the visual candidate-ranking operator; shared settings for that pairwise comparison are summarized in the experimental setup and appendix tables.

\textbf{ViDoRe v3} \citep{vidorev3} reports NDCG@10 on visually grounded slices and evaluates the same late-interaction visual evidence ranking interface against strong retrieval baselines \citep{faysse2025colpali,dong2025mmdocir}, complementing in-stack candidate-list metrics on MultimodalQA.

\subsection{Metrics and interpreting end-task scores}
\label{sec:exp:scope}

\emph{Exact match}~(EM) is computed after the benchmark's answer normalization; \emph{F1} is word-/token-level overlap between the generated answer and references.
These scores summarize the full graph-grounded pipeline and are not direct measures of candidate-list accuracy or complete graph support coverage.
Before downstream extraction and generation, graph-linked image candidates are evaluated with \emph{Hit@1}, \emph{MRR@$k$}, and \emph{nDCG@$k$}: \emph{Hit@1} indicates whether a supporting image ranks first; \emph{MRR@$k$} is the reciprocal rank of the first relevant candidate within the top-$k$ (zero if none); \emph{nDCG@$k$} discounts relevant candidates by rank under shared relevance labels.
Following the WebQA protocol~\citep{chang2022webqa}, \emph{QA-FL} is a BARTScore-based fluency measure aligned with reference answers; \emph{QA-Acc} checks key information via keyword-style rules with handling for closed answer categories; corpus \emph{QA} averages the per-example product of QA-FL and QA-Acc.
\emph{NDCG@10} summarizes retrieval quality up to rank ten on ViDoRe~v3~\citep{vidorev3}.

Retrieval-stage metrics ask whether the visual scorer promotes relevant graph-linked images early enough under the shared downstream interface; end-task EM/F1 ask whether the full pipeline yields reference-matching answers.
Concordant trends support the routing narrative, but neither metric family alone proves the other.
We therefore read MultimodalQA gains as \emph{consistent with} improved routing when modality tags align with visual mechanisms, and treat text-tagged rows as coupling diagnostics.

Table~\ref{tab:retrieval-metrics} reports ViDoRe v3~\citep{vidorev3} NDCG@10 across six domain slices (C.S., Phar., Ind., Fin., H.R., Phys.).
The Avg.\ column averages all six slices when every slice is reported; rows with missing slices show an em dash in Avg.\ and are not compared by that aggregate.
The final row evaluates the matched ColGraphRAG visual evidence ranking setting, providing retrieval-native evidence for the same graph-linked image ranking mechanism used in the GraphRAG comparison.
We also measure graph-linked image ranking before extraction/reasoning on MultimodalQA using Hit@1, MRR@3/10, and nDCG@3; because non-visual stages are unchanged, shifts isolate the visual scorer, and MaxSim improves every reported candidate-list point estimate versus pooled ranking (Table~\ref{tab:shortlist-graph-linked}).

%
%

\begin{table}[t]
  \centering
  \small
  \renewcommand{\arraystretch}{1.15}
  \setlength{\tabcolsep}{4pt}
  \caption{ViDoRe v3~\citep{vidorev3} retrieval: NDCG@10 (\%) on six slices (C.S., Phar., Ind., Fin., H.R., Phys.).
  The \ColGraphRAG{} row is the matched late-interaction visual evidence ranking setting.
  Avg.\ is the six-slice mean when every slice is numeric; otherwise ---.
  Within each type block, \textbf{bold} and \underline{underline} mark best and second-best scores per column (excluding ---).}
  \label{tab:retrieval-metrics}
  \begin{tabular}{@{}llcccccccc@{}}
    \toprule
    Type & Model & Size & C.S. & Phar. & Ind. & Fin. & H.R. & Phys. & Avg. \\
    \midrule
    \multirow{4}{*}{Textual Retrievers}
      & BGE-M3~\citep{chen2024bge_m3} & 0.6B & 58.0 & 52.0 & 28.5 & 39.8 & 42.4 & 35.9 & 42.8 \\
     & Jina-v4 (Textual)~\citep{gunther2025jinaembeddingsv4} & 3B & 64.3 & 54.9 & 38.4 & \underline{48.4} & \textbf{52.8} & \underline{43.6} & \underline{50.4} \\
     & Qwen3~\citep{qwen3embedding} & 8B & \underline{71.7} & \underline{59.2} & \underline{40.4} & \textbf{49.4} & \underline{47.6} & \textbf{45.6} & \textbf{52.3} \\
     & Qwen3.5~\citep{qwen3embedding} & 8B & \textbf{73.7} & \textbf{67.0} & \textbf{42.6} & --- & --- & --- & --- \\
    \midrule
    \multirow{9}{*}{Visual Retrievers}
      & Nomic~\citep{nomicembedmultimodal2025} & 7B & 66.6 & 58.9 & 37.9 & 48.8 & 46.2 & 44.2 & 50.4 \\
     & ColQwen2.5~\citep{faysse2025colpali} & 3B & 72.3 & 57.9 & 41.3 & 52.3 & 51.2 & 45.9 & 53.5 \\
     & ColEmbed~\citep{xu2025llamanemoretrievercolembedtopperforming} & 1B & 71.3 & 62.6 & 46.6 & 58.9 & 57.0 & 44.1 & 56.8 \\
     & ColEmbed~\citep{xu2025llamanemoretrievercolembedtopperforming} & 3B & 75.2 & 63.7 & 47.1 & 60.9 & 58.7 & 45.1 & 58.5 \\
     & ColNomic~\citep{nomicembedmultimodal2025} & 3B & 72.7 & 61.1 & 47.4 & 56.3 & 57.3 & 47.5 & 57.1 \\
     & ColNomic~\citep{nomicembedmultimodal2025} & 7B & 76.2 & 62.3 & 50.1 & 56.6 & 58.7 & \underline{48.3} & 58.7 \\
     & Jina-v4 (Visual)~\citep{gunther2025jinaembeddingsv4} & 3B & 71.8 & 63.1 & 50.4 & 59.3 & 59.5 & 46.6 & 58.5 \\
     & ColEmbed-v2~\citep{moreira2026nemotroncolembedv2} & 3B & \underline{77.1} & \underline{66.0} & \underline{51.7} & \underline{64.2} & \textbf{62.3} & 47.0 & \underline{61.4} \\
     & ColGraphRAG (Ours) & 3B & \textbf{80.2} & \textbf{74.8} & \textbf{61.9} & \textbf{79.3} & \underline{60.2} & \textbf{64.9} & \textbf{70.2} \\
    \bottomrule
  \end{tabular}
\end{table}

\subsection{Aggregate and modality-wise downstream effects}
\label{sec:exp:main}

We next ask whether improved evidence routing translates to downstream answer quality; Table~\ref{tab:mmqa-webqa-combined} reports aggregate MultimodalQA F1/EM together with paired WebQA QA-FL, QA-Acc, and QA.

Literature rows are protocol-contextual and not a joint leaderboard with our bottom comparison rows.
The bottom comparison block shares snapshot, image budget, expansion, non-visual retrievers, extraction templates, and generator---only visual ranking differs; cell emphasis follows Table~\ref{tab:mmqa-webqa-combined}'s caption (per Type block and metric column).

With the same pipeline except for the visual ranking operator, \ColGraphRAG{} improves aggregate F1/EM point estimates relative to the baseline graph-grounded row~\citep{bu2025querydriven}, matching the expectation that better \emph{candidate ranking over graph-linked image evidence} can raise end-task scores when the unchanged pipeline can exploit richer image-linked context.

Table~\ref{tab:modality-qd-colgraphrag-em-f1} decomposes \emph{MultimodalQA} F1 and EM by dominant evidence modality under the same comparison.
Image- and table-tagged questions improve in both F1 and EM, which aligns with a mechanism story in which graph-linked assets tied to visuals or to visually anchored layouts are more likely to be promoted into the ordered top-$k$ candidate list $E_k(q)$ and thus participate in extraction and subgraph formation.
The \textbf{text} rows show a slight drop in both F1 and EM under the same comparison.
Noise, coupled extraction effects when $E_k(q)$ shifts, and imperfect modality isolation can explain the dip; we treat it as diagnostic rather than a primary claim, and we expect benefits to concentrate where fine-grained query--image alignment gates inclusion whereas text/table-heavy items with incidental images should show neutral or mixed movement, consistent with the modality breakdown.
Definitions for the WebQA columns follow the preceding notation and the WebQA protocol~\citep{chang2022webqa}; literature rows in the WebQA block are contextual, while \ColGraphRAG{} uses the same metric code paths as the bottom comparison rows.

\paragraph{Interpreting the evidence-inclusion chain.}
\(E_k(q)\) gates graph-linked image evidence into \(R(q)\) and \(G^\star(q)\); ranking gains matter only when promoted assets survive evidence assembly.
We read retrieval-through-end-task scores as evidence-inclusion signals between candidate lists and QA outputs, while finer tabulations (gold retention through extraction, path coverage) remain future work.
$E_k(q)$ highlights the contrast between pooled ranking and MaxSim at the \emph{scoring interface}: pooling collapses each candidate to one global vector before matching, whereas late-interaction MaxSim preserves local units so different query tokens can align with different regions or patches, which targets omission of a graph-linked image from $E_k(q)$ even when it is present in $G$.
MaxSim promotions yield improved downstream inputs without changing $G$ or $B(q)$, whereas text- or table-dominated gold evidence yields weaker or mixed movement because editing only $E_k(q)$ may perturb coupled extraction (cf.\ text rows above); late interaction sharpens promotion into structured inputs rather than the full retrieval stack, and qualitative comparisons plus ablations over $k$, pools, and phrase templates are deferred.

\FloatBarrier
\subsection{Retrieval-native diagnostics on ViDoRe v3}
\label{sec:exp:vidore}

Table~\ref{tab:retrieval-metrics} evaluates \ColGraphRAG{}'s late-interaction visual evidence ranking on ViDoRe v3~\citep{vidorev3}.
On the matched slices, this setting achieves the strongest six-slice average NDCG@10 and large gains on Finance and Physics, while H.R.\ remains a mixed domain slice.
These results provide retrieval-native evidence that MaxSim ranks visually grounded candidates effectively, and they align with the MultimodalQA comparison in which \(G\), non-visual retrieval, extraction, expansion, and generation are shared across the paired comparison.
Appendix~\ref{app:vidore-topk-curves} supplements this aggregate slice summary with top-\(k\) retrieval curves on Physics and Finance.

\section{Conclusion}
\label{sec:conclusion}

We isolated \emph{graph-linked image candidate ranking} within a graph-grounded multimodal QA pipeline by keeping offline construction of $G$, baseline text- and table-side retrieval, and graph-grounded answering unchanged while replacing only the graph-linked image ranker with late-interaction MaxSim scoring~\cite{khattab2020colbert,faysse2025colpali}; on MultimodalQA this is associated with improved retrieval-stage candidate-ranking point estimates and higher aggregate EM/F1 point estimates, but yields mixed modality trends, including a small regression on text-dominant items.
We scope claims to that substitution without statistical significance, leaderboard optimality, or broader transfer guarantees.
Improved document QA may support education, accessibility, and analyst workflows, whereas the same retrieval machinery can facilitate large-scale extraction from sensitive corpora if deployed without safeguards.

\clearpage
\FloatBarrier
\appendix

\section{Candidate-ranking and end-task breakdowns}
\label{app:controlled-breakdown}

This section collects evaluation results on \emph{MultimodalQA} for the main-text visual-ranker comparison.
Table~\ref{tab:shortlist-graph-linked} reports retrieval-stage quality of graph-linked image candidate ranking; Table~\ref{tab:modality-qd-colgraphrag-em-f1} reports end-to-end F1 and EM by dominant evidence modality on the same benchmark.


\begin{table}[H]
  \nolinenumbers
  \centering
  \small
  \caption{\textbf{MultimodalQA.} Retrieval-stage quality when replacing pooled visual ranking with late-interaction MaxSim ranking, measured on ranked candidates from $\mathcal{X}_{\mathrm{img}}(G)$ \emph{before} downstream graph extraction and generation.
  The pooled single-vector variant uses global image representations, whereas the late-interaction MaxSim variant ranks candidates in $\mathcal{X}_{\mathrm{img}}(G)$ with token/patch-level matching.}
  \label{tab:shortlist-graph-linked}
  \setlength{\tabcolsep}{6pt}
  \begin{tabular*}{\textwidth}{@{\extracolsep{\fill}}lcccc@{}}
    \toprule
    Method & Hit@1 & MRR@3 & MRR@10 & nDCG@3 \\
    \midrule
    Pooled single-vector ranking & \underline{38.0} & \underline{61.0} & \underline{62.5} & \underline{32.6} \\
    Late-interaction MaxSim ranking & \textbf{57.0} & \textbf{78.5} & \textbf{78.5} & \textbf{39.5} \\
    \bottomrule
  \end{tabular*}
  \linenumbers
\end{table}


\begin{table}[H]
  \nolinenumbers
  \centering
  \caption{\textbf{MultimodalQA.} End-to-end diagnostic breakdown when replacing pooled visual ranking with late-interaction MaxSim ranking. F1 and EM are reported by dominant gold evidence modality (image, table, text), matching Table~\ref{tab:mmqa-webqa-combined}.}
  \label{tab:modality-qd-colgraphrag-em-f1}
  \small
  \setlength{\tabcolsep}{5pt}
  \begin{tabular*}{\textwidth}{@{\extracolsep{\fill}}lcccccc@{}}
    \toprule
    \multirow{2}{*}{Method}
      & \multicolumn{2}{c}{image}
      & \multicolumn{2}{c}{table}
      & \multicolumn{2}{c@{}}{text} \\
    \cmidrule(lr){2-3} \cmidrule(lr){4-5} \cmidrule(l){6-7}
     & F1 & EM & F1 & EM & F1 & EM \\
    \midrule
    Pooled single-vector ranking & \underline{18.3} & \underline{16.0} & \underline{73.4} & \underline{71.8} & \textbf{59.7} & \textbf{58.3} \\
    Late-interaction MaxSim ranking & \textbf{24.5} & \textbf{20.0} & \textbf{80.2} & \textbf{76.9} & \underline{57.9} & \underline{55.6} \\
    \bottomrule
  \end{tabular*}
  \linenumbers
\end{table}

\section{ViDoRe v3 subsets: retrieval performance across top-$k$}
\label{app:vidore-topk-curves}

\begin{figure}[H]
  \centering
  \includegraphics[width=0.43\linewidth]{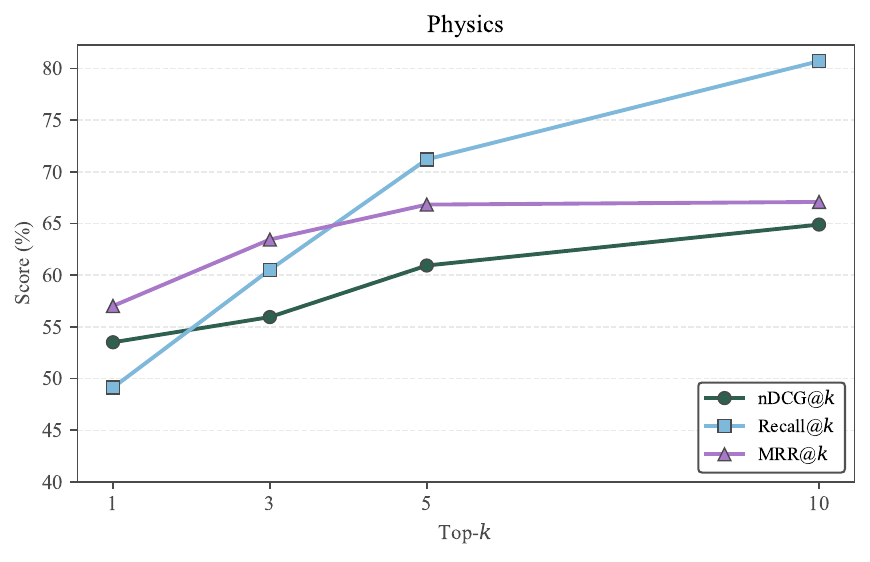}\hfill
  \includegraphics[width=0.43\linewidth]{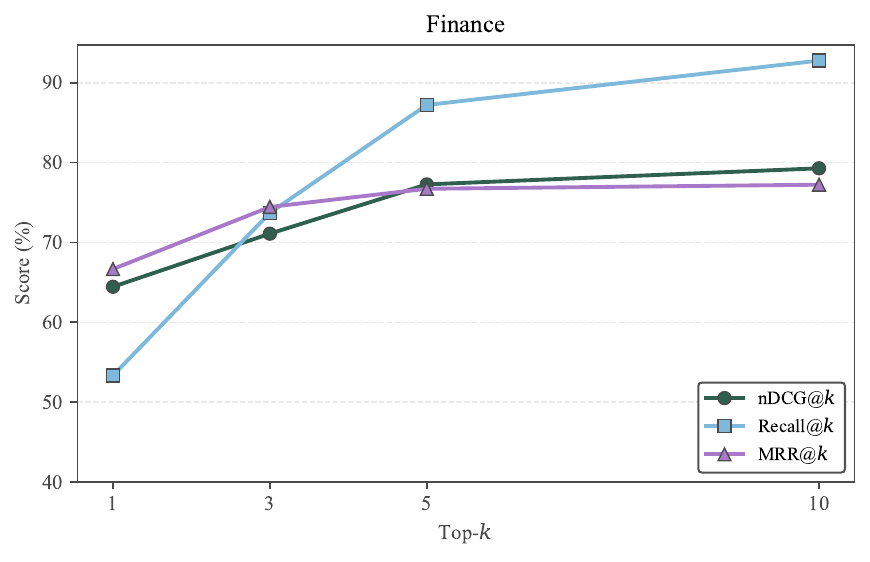}
  \caption{ViDoRe v3 subsets: retrieval performance across top-$k$.
  We report nDCG@$k$, Recall@$k$, and MRR@$k$ for the Physics (left) and Finance (right) subsets.
  Larger candidate budgets increase recall-oriented coverage; nDCG@$k$ and MRR@$k$ additionally indicate whether promoted candidates remain ranked early enough to be useful, with stronger early saturation on Finance and more gradual movement on Physics.}
  \label{fig:appendix-topk-vidore-en}
\end{figure}

Figure~\ref{fig:appendix-topk-vidore-en} complements the aggregate ViDoRe~v3 slice summary in Table~\ref{tab:retrieval-metrics}~\citep{vidorev3}.
As the candidate budget increases, Recall@$k$ improves, while nDCG@$k$ and MRR@$k$ indicate whether relevant evidence remains concentrated near the top of the ranked list.
The \emph{Physics} subset exhibits a gradual gain as $k$ grows, while the \emph{Finance} subset reaches strong retrieval quality with a smaller candidate budget and shows earlier saturation.

\section{Qualitative MultimodalQA examples}
\label{app:qualitative-mmqa}

\begin{figure}[H]
  \centering
  \includegraphics[width=\textwidth]{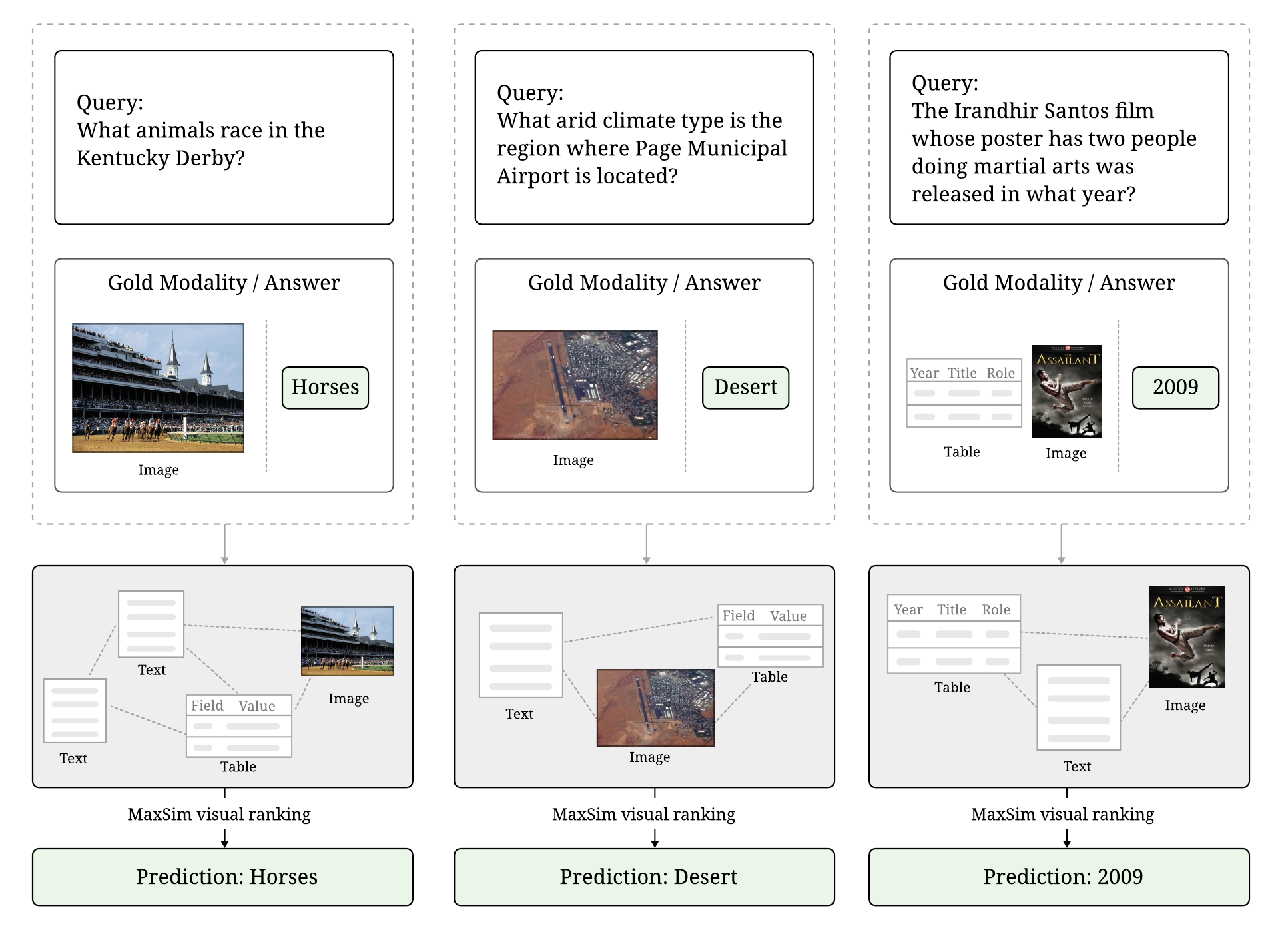}
  \caption{Qualitative MultimodalQA examples.
  Each panel shows a question, gold evidence, a compact retrieved evidence graph, and the final prediction.
  The examples illustrate how graph-linked visual evidence ranking supports image-grounded and table--image compositional reasoning in ColGraphRAG.}
  \label{fig:appendix-mmqa-qualitative}
\end{figure}

\paragraph{Qualitative examples.}
Figure~\ref{fig:appendix-mmqa-qualitative} illustrates how to read the qualitative MultimodalQA cases: each panel shows the query, gold evidence and answer, a compact retrieved evidence graph, and the prediction after MaxSim visual ranking.
The examples cover two image-grounded questions and one table--image compositional question, where the visual cue helps identify the table row used to answer.

{
\small
\begin{thebibliography}{99}

\bibitem{talmor2021multimodalqa}
A.~Talmor, O.~Yoran, A.~Catav, D.~Lahav, Y.~Wang, A.~Asai, G.~Ilharco, H.~Hajishirzi, and J.~Berant.
\newblock MultiModalQA: Complex Question Answering over Text, Tables and Images.
\newblock In \emph{International Conference on Learning Representations (ICLR)}, 2021.

\bibitem{suri2025visdom}
M.~Suri, P.~Mathur, F.~Dernoncourt, K.~Goswami, R.~A.~Rossi, and D.~Manocha.
\newblock VisDoM: Multi-Document QA with Visually Rich Elements Using Multimodal Retrieval-Augmented Generation.
\newblock In \emph{Proceedings of NAACL}, 2025.

\bibitem{tanaka2025vdocrag}
R.~Tanaka, T.~Iki, T.~Hasegawa, K.~Nishida, K.~Saito, and J.~Suzuki.
\newblock VDocRAG: Retrieval-Augmented Generation over Visually-Rich Documents.
\newblock In \emph{Proceedings of CVPR}, 2025.

\bibitem{ma2024mmlongbenchdoc}
Y.~Ma, Y.~Zang, L.~Chen, M.~Chen, Y.~Jiao, X.~Li, X.~Lu, Z.~Liu, Y.~Ma, X.~Dong, P.~Zhang, L.~Pan, Y.-G.~Jiang, J.~Wang, Y.~Cao, and A.~Sun.
\newblock MMLongBench-Doc: Benchmarking Long-context Document Understanding with Visualizations.
\newblock In \emph{Advances in Neural Information Processing Systems}, 2024.

\bibitem{deng2025longdocurl}
C.~Deng, J.~Yuan, P.~Bu, P.~Wang, Z.-Z.~Li, J.~Xu, X.-H.~Li, Y.~Gao, J.~Song, B.~Zheng, and C.-L.~Liu.
\newblock LongDocURL: a Comprehensive Multimodal Long Document Benchmark Integrating Understanding, Reasoning, and Locating.
\newblock In \emph{Proceedings of ACL (Long Papers)}, 2025.

\bibitem{yang2018hotpotqa}
Z.~Yang, P.~Qi, S.~Zhang, Y.~Bengio, W.~Cohen, R.~Salakhutdinov, and C.~D.~Manning.
\newblock HotpotQA: A Dataset for Diverse, Explainable Multi-hop Question Answering.
\newblock In \emph{Proceedings of EMNLP}, 2018.

\bibitem{trivedi2022musique}
H.~Trivedi, N.~Balasubramanian, T.~Khot, and A.~Sabharwal.
\newblock MuSiQue: Multihop Questions via Single-hop Question Composition.
\newblock \emph{Transactions of the Association for Computational Linguistics}, 10:539--554, 2022.

\bibitem{he2023multimodalgraphtransformer}
X.~He and X.~Wang.
\newblock Multimodal Graph Transformer for Multimodal Question Answering.
\newblock In \emph{Proceedings of EACL}, 2023.

\bibitem{bu2025querydriven}
C.~Bu, G.~Chang, Z.~Chen, C.~Dang, Z.~Wu, Y.~He, and X.~Wu.
\newblock Query-Driven Multimodal GraphRAG: Dynamic Local Knowledge Graph Construction for Online Reasoning.
\newblock In \emph{Findings of the Association for Computational Linguistics: ACL 2025}, pages 21360--21380, Vienna, Austria.
\newblock Association for Computational Linguistics.
\newblock DOI: \href{https://doi.org/10.18653/v1/2025.findings-acl.1100}{10.18653/v1/2025.findings-acl.1100}.
\newblock URL: \url{https://aclanthology.org/2025.findings-acl.1100/}.

\bibitem{chang2022webqa}
Y.~Chang, M.~Narang, H.~Suzuki, G.~Cao, J.~Gao, and Y.~Bisk.
\newblock WebQA: Multihop and Multimodal Question Answering.
\newblock In \emph{Proceedings of the IEEE/CVF Conference on Computer Vision and Pattern Recognition (CVPR)}, pages 16495--16504, 2022.

\bibitem{yang-etal-2025-superrag}
C.~Yang, D.-K.~Vu, M.-T.~Nguyen, X.-Q.~Nguyen, L.~Nguyen, and H.~Le.
\newblock SuperRAG: Beyond RAG with Layout-Aware Graph Modeling.
\newblock In \emph{Proceedings of NAACL (Industry Track)}, 2025.

\bibitem{radford2021clip}
A.~Radford, J.~W.~Kim, C.~Hallacy, A.~Ramesh, G.~Goh, S.~Agarwal, G.~Sastry, A.~Askell, P.~Mishkin, J.~Clark, G.~Krueger, and I.~Sutskever.
\newblock Learning Transferable Visual Models From Natural Language Supervision.
\newblock In \emph{Proceedings of ICML}, 2021.

\bibitem{khattab2020colbert}
O.~Khattab and M.~Zaharia.
\newblock ColBERT: Efficient and Effective Passage Search via Contextualized Late Interaction over BERT.
\newblock In \emph{Proceedings of SIGIR}, 2020.

\bibitem{santhanam2022colbertv2}
K.~Santhanam, O.~Khattab, J.~Saad-Falcon, C.~Potts, and M.~Zaharia.
\newblock ColBERTv2: Effective and Efficient Retrieval via Lightweight Late Interaction.
\newblock In \emph{Proceedings of NAACL}, 2022.

\bibitem{faysse2025colpali}
M.~Faysse, H.~Sibille, T.~Wu, B.~Omrani, G.~Viaud, C.~Hudelot, and P.~Colombo.
\newblock ColPali: Efficient Document Retrieval with Vision Language Models.
\newblock In \emph{International Conference on Learning Representations (ICLR)}, 2025.

\bibitem{lin2023flmr}
W.~Lin, J.~Chen, J.~Mei, A.~Coca, and B.~Byrne.
\newblock Fine-grained Late-interaction Multi-modal Retrieval for Retrieval Augmented Visual Question Answering.
\newblock In \emph{Advances in Neural Information Processing Systems}, 2023.

\bibitem{lin2024preflmr}
W.~Lin, J.~Mei, J.~Chen, and B.~Byrne.
\newblock PreFLMR: Scaling Up Fine-Grained Late-Interaction Multi-modal Retrievers.
\newblock In \emph{Proceedings of the 62nd Annual Meeting of the Association for Computational Linguistics (Volume 1: Long Papers)}, 2024.

\bibitem{dong2025mmdocir}
K.~Dong, Y.~Chang, D.~Goh, D.~Li, R.~Tang, and Y.~Liu.
\newblock MMDocIR: Benchmarking Multimodal Retrieval for Long Documents.
\newblock In \emph{Proceedings of the 2025 Conference on Empirical Methods in Natural Language Processing}, 2025.

\bibitem{wasserman2025real}
N.~Wasserman, R.~Pony, O.~Naparstek, A.~R.~Goldfarb, E.~Schwartz, U.~Barzelay, and L.~Karlinsky.
\newblock REAL-MM-RAG: A Real-World Multi-Modal Retrieval Benchmark.
\newblock In \emph{Proceedings of the 63rd Annual Meeting of the Association for Computational Linguistics (Volume 1: Long Papers)}, 2025.

\bibitem{chia2025mlongdoc}
Y.~K.~Chia, L.~Cheng, H.~P.~Chan, M.~Song, C.~Liu, M.~Aljunied, S.~Poria, and L.~Bing.
\newblock M-LongDoc: A Benchmark For Multimodal Super-Long Document Understanding And A Retrieval-Aware Tuning Framework.
\newblock In \emph{Proceedings of the 2025 Conference on Empirical Methods in Natural Language Processing}, 2025.

\bibitem{jain2025simpledoc}
C.~Jain, Y.~Wu, Y.~Zeng, J.~Liu, S.~Dai, Z.~Shao, Q.~Wu, and H.~Wang.
\newblock SimpleDoc: Multi-Modal Document Understanding with Dual-Cue Page Retrieval and Iterative Refinement.
\newblock In \emph{Proceedings of the 2025 Conference on Empirical Methods in Natural Language Processing}, 2025.

\bibitem{vanlandeghem2023dude}
J.~Van Landeghem, R.~Tito, {\L}.~Borchmann, M.~Pietruszka, P.~Joziak, R.~Powalski, D.~Jurkiewicz, M.~Coustaty, B.~Anckaert, E.~Valveny, M.~Blaschko, S.~Moens, and T.~Stanislawek.
\newblock Document Understanding Dataset and Evaluation (DUDE).
\newblock In \emph{Proceedings of the IEEE/CVF International Conference on Computer Vision (ICCV)}, 2023.

\bibitem{blau2024gram}
T.~Blau, S.~Fogel, R.~Ronen, A.~Golts, R.~Ganz, E.~Ben Avraham, A.~Aberdam, S.~Tsiper, and R.~Litman.
\newblock GRAM: Global Reasoning for Multi-Page VQA.
\newblock In \emph{Proceedings of the IEEE/CVF Conference on Computer Vision and Pattern Recognition (CVPR)}, 2024.

\bibitem{chen2024bge_m3}
J.~Chen, S.~Xiao, P.~Zhang, K.~Luo, D.~Lian, and Z.~Liu.
\newblock {BGE} {M3}-Embedding: Multi-lingual, multi-functionality, multi-granularity text embeddings through self-knowledge distillation.
\newblock \emph{arXiv preprint arXiv:2402.03216}, 2024.

\bibitem{gunther2025jinaembeddingsv4}
M.~G{\"u}nther, S.~Sturua, M.~K.~Akram, I.~Mohr, A.~Ungureanu, S.~Eslami, S.~Martens, B.~Wang, N.~Wang, and H.~Xiao.
\newblock jina-embeddings-v4: Universal embeddings for multimodal multilingual retrieval.
\newblock \emph{arXiv preprint arXiv:2506.18902}, 2025.

\bibitem{qwen3embedding}
Y.~Zhang, M.~Li, D.~Long, X.~Zhang, H.~Lin, B.~Yang, P.~Xie, A.~Yang, D.~Liu, J.~Lin, F.~Huang, and J.~Zhou.
\newblock Qwen3 Embedding: Advancing text embedding and reranking through foundation models.
\newblock \emph{arXiv preprint arXiv:2506.05176}, 2025.

\bibitem{qwen25vl}
S.~Bai, K.~Chen, X.~Liu, J.~Wang, W.~Ge, S.~Song, K.~Dang, P.~Wang, S.~Wang, J.~Tang, et al.
\newblock Qwen2.5-VL technical report.
\newblock \emph{arXiv preprint arXiv:2502.13923}, 2025.

\bibitem{nomicembedmultimodal2025}
{Nomic Team}.
\newblock Nomic Embed Multimodal: Interleaved text, image, and screenshots for visual document retrieval.
\newblock Nomic~AI blog, 2025.
\newblock Available at \url{https://nomic.ai/blog/posts/nomic-embed-multimodal}.

\bibitem{xu2025llamanemoretrievercolembedtopperforming}
M.~Xu, G.~Moreira, R.~Ak, R.~Osmulski, Y.~Babakhin, Z.~Yu, B.~Schifferer, and E.~Oldridge.
\newblock Llama Nemoretriever Colembed: Top-performing text-image retrieval model.
\newblock \emph{arXiv preprint arXiv:2507.05513}, 2025.

\bibitem{moreira2026nemotroncolembedv2}
G.~d.~S.~P.~Moreira, R.~Ak, M.~Xu, O.~Holworthy, B.~Schifferer, Z.~Yu, Y.~Babakhin, R.~Osmulski, J.~Cai, R.~Chesler, B.~Liu, and E.~Oldridge.
\newblock Nemotron {ColEmbed V2}: Top-performing late interaction embedding models for visual document retrieval.
\newblock \emph{arXiv preprint arXiv:2602.03992}, 2026.

\bibitem{liu2023mmhqa_icl}
W.~Liu, F.~Lei, T.~Luo, J.~Lei, S.~He, J.~Zhao, and K.~Liu.
\newblock {MMHQA-ICL}: Multimodal in-context learning for hybrid question answering over text, tables and images.
\newblock \emph{arXiv preprint arXiv:2309.04790}, 2023.

\bibitem{chen2022murag}
W.~Chen, H.~Hu, X.~Chen, P.~Verga, and W.~Cohen.
\newblock {MuRAG}: Multimodal retrieval-augmented generator for open question answering over images and text.
\newblock In \emph{Proceedings of the Conference on Empirical Methods in Natural Language Processing}, pages 5558--5570, 2022.

\bibitem{yang2023enhancing_multimodal_multihop}
Q.~Yang, Q.~Chen, W.~Wang, B.~Hu, and M.~Zhang.
\newblock Enhancing multi-modal multi-hop question answering via structured knowledge and unified retrieval-generation.
\newblock In \emph{Proceedings of the 31st ACM International Conference on Multimedia}, pages 5223--5234, 2023.

\bibitem{chiang2023vicuna}
W.-L.~Chiang, Z.~Li, Z.~Lin, Y.~Sheng, Z.~Wu, H.~Zhang, L.~Zheng, S.~Zhuang, Y.~Zhuang, J.~E.~Gonzalez, I.~Stoica, and E.~P.~Xing.
\newblock Vicuna: An open-source chatbot impressing {GPT-4} with 90\%* {ChatGPT} quality.
\newblock {LMSYS} Blog, 2023.
\newblock Available at \url{https://lmsys.org/blog/2023-03-30-vicuna/}.

\bibitem{touvron2023llama2}
H.~Touvron, L.~Martin, K.~Stone, P.~Albert, A.~Almahairi, Y.~Babaei, N.~Bashlykov, S.~Batra, P.~Bhargava, S.~Bhosale, et al.
\newblock {Llama} 2: Open foundation and fine-tuned chat models.
\newblock \emph{arXiv preprint arXiv:2307.09288}, 2023.

\bibitem{wang2024openchat}
G.~Wang, S.~Cheng, X.~Zhan, X.~Li, S.~Song, and Y.~Liu.
\newblock {OpenChat}: Advancing open-source language models with mixed-quality data.
\newblock In \emph{International Conference on Learning Representations}, 2024.

\bibitem{zhang2023moqagpt}
L.~Zhang, Y.~Wu, F.~Mo, J.-Y.~Nie, and A.~Agrawal.
\newblock {MoqaGPT}: Zero-shot multi-modal open-domain question answering with large language model.
\newblock In \emph{Findings of the Association for Computational Linguistics: EMNLP}, pages 1195--1210, 2023.

\bibitem{wang2022ofa}
P.~Wang, A.~Yang, R.~Men, J.~Lin, S.~Bai, Z.~Li, J.~Ma, C.~Zhou, J.~Zhou, and H.~Yang.
\newblock {OFA}: Unifying architectures, tasks, and modalities through a simple sequence-to-sequence learning framework.
\newblock In \emph{Proceedings of the 39th International Conference on Machine Learning}, pages 23318--23340, 2022.

\bibitem{hu2023promptcap}
Y.~Hu, H.~Hua, Z.~Yang, W.~Shi, N.~A.~Smith, and J.~Luo.
\newblock {PromptCap}: Prompt-guided image captioning for {VQA} with {GPT-3}.
\newblock In \emph{Proceedings of the IEEE/CVF International Conference on Computer Vision (ICCV)}, pages 2963--2975, 2023.

\bibitem{cheng2023binding}
Z.~Cheng, T.~Xie, P.~Shi, C.~Li, R.~Nadkarni, Y.~Hu, C.~Xiong, D.~Radev, M.~Ostendorf, L.~Zettlemoyer, N.~A.~Smith, and T.~Yu.
\newblock Binding language models in symbolic languages.
\newblock In \emph{International Conference on Learning Representations}, 2023.

\bibitem{vidorev3}
A.~Loison, Q.~Mac{\'e}, A.~Edy, V.~Xing, T.~Balough, G.~Moreira, B.~Liu, M.~Faysse, C.~Hudelot, and G.~Viaud.
\newblock {ViDoRe V3}: A comprehensive evaluation of retrieval augmented generation in complex real-world scenarios.
\newblock \emph{arXiv preprint arXiv:2601.08620}, 2026.

\end{thebibliography}
}
\end{document}